\DocumentMetadata{}
\PassOptionsToPackage{table,xcdraw}{xcolor}

\documentclass[sigconf, review=false]{acmart}

\usepackage{booktabs}
\usepackage[T1]{fontenc}
\usepackage{tabularray}
\usepackage{tabularx}
\usepackage{pifont}
\usepackage{amsmath}
\usepackage{amsfonts}
\usepackage{graphicx}
\usepackage{subcaption}
\usepackage{caption}
\usepackage{comment}
\usepackage{multirow}
\usepackage{url}

\usepackage{notoccite}
\usepackage{hyperref}
\usepackage{stfloats}

\copyrightyear{2025}
\acmYear{2025}
\setcopyright{acmcopyright}
\acmConference[ICVGIP 2025]{Indian Conference on Computer Vision, Graphics, and Image Processing}{December 17--20, 2025}{Mandi, India}
\acmBooktitle{Indian Conference on Computer Vision, Graphics, and Image Processing (ICVGIP 2025), December 17--20, 2025, Mandi, India}
\acmPrice{}
\acmDOI{10.1145/3774521.3774577}
\acmISBN{979-8-4007-1930-1/2025/12}
\def\etal{\emph{et al}.}

\begin{document}

\title{AQIFormer: A Transformer-Based Multi-View Architecture for Cross-City Air Quality Classification}

\author{Om Kathalkar}
\affiliation{%
  \institution{IIIT Hyderabad}
  \country{India}
  }
\email{om.kathalkar@research.iiit.ac.in}

\author{Nitin Nilesh}
\affiliation{%
  \institution{IIIT Hyderabad}
  \country{India}
  }
\email{nitin.nilesh@alumni.iiit.ac.in}

\author{Sachin Chaudhari}
\affiliation{%
  \institution{IIIT Hyderabad}
  \country{India}
  }
\email{sachin.chaudhari@iiit.ac.in}

\author{Anoop Namboodiri}
\affiliation{%
  \institution{IIIT Hyderabad}
  \country{India}
  }
\email{anoop@iiit.ac.in}

\renewcommand{\shortauthors}{}

\begin{abstract}
Air pollution represents one of the most critical environmental and public health challenges globally, with traditional sensor-based monitoring systems facing significant scalability and economic constraints. Image-based air quality estimation has emerged as a promising alternative, leveraging the visual characteristics of atmospheric pollutants in traffic scenes. However, existing methods suffer from limited cross-city generalization and inadequate exploitation of multi-view perspectives. We present AQIFormer, a novel transformer-based ensemble architecture that addresses these fundamental limitations through innovative dual-view integration, weather-aware attention mechanisms, and comprehensive multi-task learning. Our approach uniquely combines front and rear traffic imagery with meteorological parameters to achieve robust air quality classification across diverse urban environments. Extensive evaluation on a comprehensive dataset of 26,678 synchronized front-rear image pairs demonstrates good performance with 89.96\% accuracy, representing a 14.96\% improvement over state-of-the-art methods. Most importantly, our model maintains exceptional cross-city generalization capabilities, achieving 81.67\% accuracy on an independent dataset collected in Nagpur, India with only 8.29\% performance degradation using few-shot adaptation with minimal training samples. \textcolor{blue}{\textbf{\href{https://india-data.org/dataset-details/e0f80763-0243-4fc2-8a56-0537824cbd9a}{Dataset Link}}}
\end{abstract}




\maketitle

\section{Introduction}

Air pollution represents one of the most critical environmental and public health challenges globally, contributing to approximately 7 million premature deaths annually worldwide~\cite{who}. This crisis primarily results through elevated concentrations of fine particulate matter (PM2.5 and PM10) in the atmosphere, with vehicular emissions emerging as the dominant contributor in urban environments~\cite{zhang2013air}.

India faces particularly severe air quality challenges, with 21 of the world's 30 most polluted cities located within its borders \cite{cnn2020india}. The country's rapid urbanization and exponential vehicular growth from 55 million in 2001 to over 295 million in 2019~\cite{Goel2015VehicularGrowth} have created a complex pollution landscape where PM2.5 and PM10 concentrations frequently exceed WHO guidelines by factors of 5-10~\cite{guttikunda2019air}. Traffic-related emissions account for approximately 40-50\% of urban air pollution, creating unique monitoring challenges due to their highly localized and dynamic nature. Vehicle emissions generate pollution hotspots along major corridors, with concentration gradients varying dramatically within hundreds of meters~\cite{Zhu2002Concentration}. Traditional monitoring approaches fail to capture these micro-scale variations, as pollutant levels can differ by orders of magnitude between busy intersections and nearby residential areas, while temporal dynamics further complicate efforts with concentrations fluctuating rapidly based on traffic density, vehicle types, and meteorological conditions.

The Air Quality Index~(AQI) serves as a standardized metric for communicating air quality conditions to the public, ranging from 0 to 500, where lower values indicate healthier air conditions. The Central Pollution Control Board (CPCB) of India has established six AQI categories: Good (0-50), Satisfactory (51-100), Moderate (101-200), Poor (201-300), Very Poor (301-400), and Severe (>400)~\cite{cpcb2014aqi}. Traditional AQI estimation relies on sophisticated sensor networks that measure PM concentrations and compute composite indices using established mathematical formulations.

Conventional air quality monitoring infrastructure faces significant scalability constraints that limit comprehensive urban coverage. India operates approximately 800 monitoring stations across 4,000+ cities, resulting in sparse spatial coverage inadequate for capturing localized traffic-related pollution variations~\cite{parmar2024development}. High-precision monitoring stations require substantial investment, with individual units costing INR 40 lakhs to 1.6 crores \cite{perfectpollucon2024}, excluding installation and maintenance expenses. Low-cost sensor alternatives, while substantially reducing initial deployment costs, introduce significant reliability and maintenance challenges that severely constrain scalability for large-scale implementation. Popular devices such as the widely used Nova SDS011 PM sensor exhibit severely limited operating lifetimes of approximately 8,000 hours, equivalent to just one year of continuous operation, requiring frequent replacement and re-calibration~\cite{nilesh2022iot}.

\subsection{Related Work}

As sensor-based methods face significant scalability issues, research has emerged toward image-based approaches that leverage the visual effects of atmospheric pollutants. Liu \etal~\cite{liu2016particle} pioneered the systematic application of light attenuation and color information for particulate matter~(PM) estimation, demonstrating that pollution levels can be inferred from observable changes in image characteristics. This approach exploits the physical principle that airborne particles scatter light, creating measurable effects on image brightness, contrast, and color saturation that correlate with air quality conditions.

The introduction of deep learning marked a significant advancement in image-based air quality estimation. Chakma \etal~\cite{chakma2017air} pioneered the use of deep CNNs for AQI classification, employing fine-tuned VGG19 architecture and achieving the classification accuracy of 68. 74\% in Beijing datasets. This work established the feasibility of using deep neural networks to extract pollution-related characteristics from atmospheric imagery. Building upon these foundations, Zhang \etal~\cite{zhang2020deep} proposed AQC-Net, incorporating Spatial and Context Attention (SCA) blocks with ResNet architecture, achieving good performance on standard datasets. The SCA mechanism enabled the model to focus on pollution-relevant regions while suppressing irrelevant background information.

Further advancing real-time capabilities, Zhang \etal~\cite{zhang2022real} developed YOLO-AQI, adapting object detection architectures for rapid AQI classification with 75.15\% accuracy and 0.058 seconds per image inference time. Recognition of single-modality limitations drove research toward multi-modal fusion strategies that combine visual information with meteorological parameters. Kalajdjieski \etal~\cite{kalajdjieski2020air} developed influential frameworks combining InceptionV3 for image feature extraction with multi-layer perceptrons for weather data processing, demonstrating significant improvements when meteorological context is incorporated. Kamble and Champrasert~\cite{kamble2021using} proposed sophisticated multi-modal frameworks that combine CNN-extracted visual features with meteorological data through advanced fusion techniques.

Sequential modeling approaches emerged to leverage temporal consistency in air quality patterns. Song \etal~\cite{song2020resnet} developed ResNet-LSTM architectures for processing smartphone image sequences, demonstrating that temporal context significantly improves classification performance by capturing the gradual evolution of atmospheric conditions. Wang \etal~\cite{wang2024surveillance} proposed CNN-LSTM hybrid models for surveillance camera feeds, achieving 94\% accuracy by exploiting sequential data continuity. Nilesh \etal~\cite{nilesh2022iot} developed IoT-based approaches combining image processing with machine learning methods, utilizing vehicle detection and visibility metrics alongside weather parameters for AQI estimation, highlighting the importance of incorporating traffic-specific features for air quality assessment.

Regional implementations have highlighted significant challenges in cross-city generalization. Mondal \etal~\cite{mondal2024uncovering} utilized smartphone captured images with custom CNNs for PM2.5 prediction in Dhaka, achieving promising localized results but demonstrating limited generalization to other urban environments. Kow \etal~\cite{kow2022real} introduced CNN-RC for multiple air quality metrics in Kaohsiung, Taiwan, while Ahmed \etal~\cite{ahmed2022aqe} developed AQE-Net for Karachi environments. These studies consistently showed strong performance within their target cities but limited transferability across different urban characteristics, highlighting a fundamental challenge in the field that stems from variations in urban layout, traffic patterns, climatic conditions, and pollution source distributions.

The transformer revolution has fundamentally transformed numerous domains through self-attention mechanisms, yet their application to environmental monitoring remains relatively unexplored. The introduction of Vision Transformers (ViTs)~\cite{dosovitskiy2020image} opened new possibilities for sophisticated spatial reasoning in computer vision tasks. Li \etal~\cite{li2023deep} demonstrated transformers' superiority over CNN-LSTM architectures for air quality prediction, though their focus remained on sensor-based time series rather than image-based approaches. Despite the proven effectiveness of transformers in various computer vision tasks, their specific application to image-based AQI classification remains largely unexplored, representing a significant gap that motivates our work.

\subsection{Contributions}

We present AQIFormer, a transformer-based multi-view architecture for cross-city air quality classification using traffic imagery. Our key contributions are:

1) A novel transformer-based architecture that integrates dual traffic imagery with weather-aware attention mechanisms, achieving 11.46\% improvement over single-view approaches through adaptive environmental processing. The learnable attention-based fusion of front and rear camera perspectives enables comprehensive scene understanding for robust air quality assessment across diverse urban environments.

2) Exceptional cross-city generalization capabilities with 81.67\% accuracy on independent city data and only 8.29\% performance degradation using few-shot adaptation, significantly outperforming existing methods that exhibit 30\%+ degradation. This establishes practical viability for scalable deployment across diverse urban environments with minimal data requirements, addressing a fundamental challenge in environmental monitoring systems.

3) A comprehensive evaluation framework incorporating multi-task learning that improves feature discriminability by 11.57\% and attention visualization demonstrating physically meaningful focus on pollution sources. We will release the Nagpur dataset to enable further cross-city research, while Table~\ref{tab:ablation} validates the contribution of each architectural component and Figure~\ref{fig:attention_maps} showcases the interpretability across various AQI categories.

The remainder of this paper is organized as follows. Section II presents our methodology, including problem formulation, AQIFormer architecture with dual-view integration and weather-aware attention mechanisms, and multi-task learning framework. Section III details the experimental setup, including TRAQID and Nagpur dataset characteristics, implementation configuration, and evaluation metrics. Section IV presents comprehensive results and analysis, covering within-city performance evaluation, cross-city generalization capabilities, ablation studies, seasonal performance analysis, and practical deployment considerations. Section V concludes with contributions summary and establishes the significance of transformer-based approaches for scalable urban air quality monitoring.
% ############################################################################################################################################################################
\section{Methodology}
This section presents AQIFormer, our transformer-based architecture for cross-city air quality classification using traffic imagery. We begin by formulating the problem as a multi-class classification task with auxiliary learning objectives. Our approach integrates dual-view traffic images with meteorological parameters through a novel fusion mechanism, followed by an enhanced transformer architecture that incorporates weather-aware attention and temporal encoding. 
\subsection{Problem Formulation and Theoretical Framework}

We formulate the air quality classification task as a multi-class classification problem enhanced with auxiliary task learning to improve feature representation and model robustness. Given synchronized front and rear traffic images $(I_f, I_r) \in \mathbb{R}^{H \times W \times C}$ and associated meteorological parameters $W = \{T, H, S, D\}$ representing temperature, humidity, season, and time-of-day respectively, we aim to learn a comprehensive mapping function:

$$f: (I_f, I_r, W) \rightarrow \{C_1, C_2, C_3, C_4, C_5, C_6\}$$

\noindent where $C_i$ represents the six AQI categories established by the Central Pollution Control Board (CPCB) standards: Good (0-50), Satisfactory (51-100), Moderate (101-200), Poor (201-300), Very Poor (301-400), and Severe (>400).

To enhance the learning process and improve feature discriminability, we incorporate auxiliary tasks that leverage the strong correlations between temporal patterns and air quality dynamics:

$$\mathcal{T}_{aux} = \{\mathcal{T}_{time}, \mathcal{T}_{season}\}$$

\noindent where $\mathcal{T}_{time}$ represents time-of-day classification (day/night) and $\mathcal{T}_{season}$ represents seasonal classification (summer/monsoon/winter). These auxiliary tasks provide additional supervision signals that guide the model to learn more discriminative features while capturing the underlying temporal patterns that influence air quality conditions.

\begin{figure*}[t]
\centering
\includegraphics[width=\textwidth]{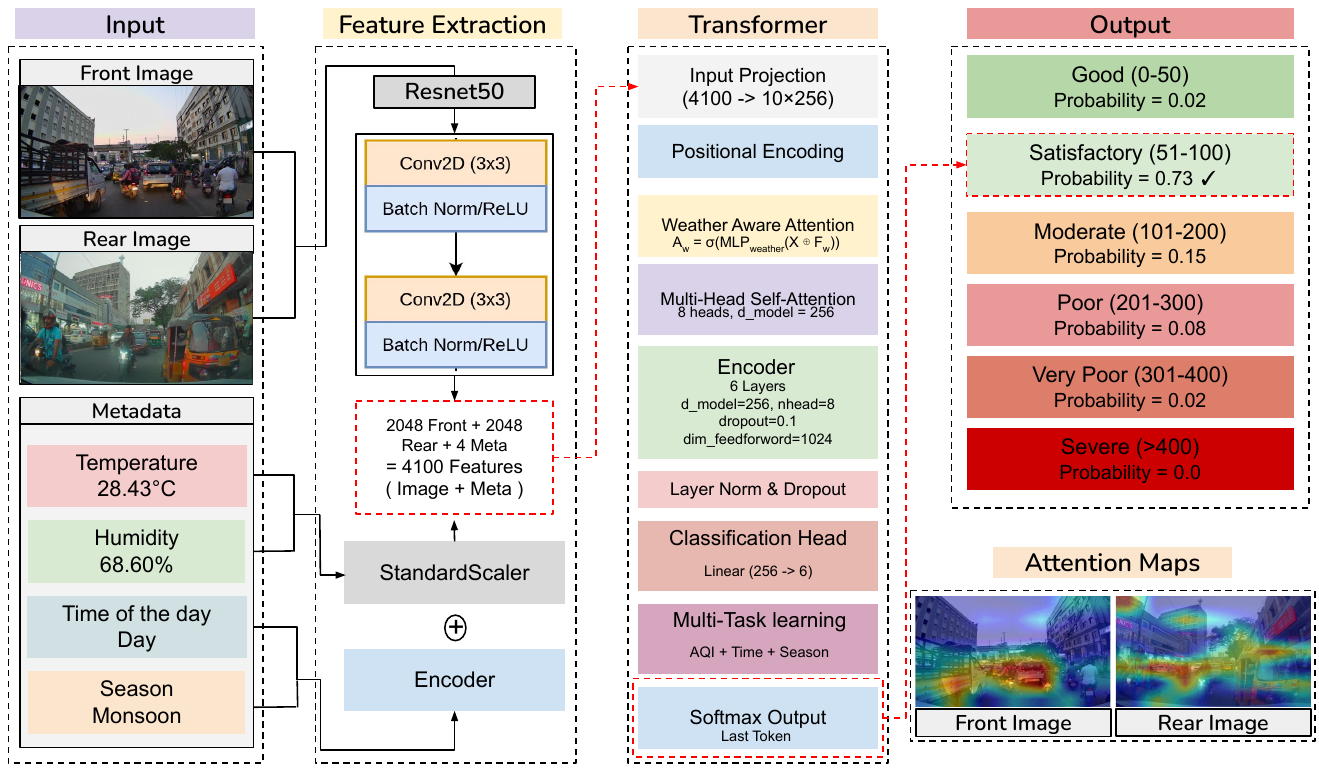}
\caption{AQIFormer architecture for multi-view air quality classification using dual traffic imagery, meteorological data, and transformer-based processing to predict AQI categories.}
\label{fig:arch}
\Description{arch}
\end{figure*}

\subsection{AQIFormer Architecture Overview}

AQIFormer consists of five main components that work together to classify air quality from traffic images. The architecture addresses the main challenges in image-based air quality estimation while ensuring the model can generalize across different cities.

The feature extraction component uses pre-trained ResNet50 networks to extract visual features from both front and rear traffic images. Meteorological parameters are processed through standardization and encoding before being directly concatenated with visual features. The dual-view integration mechanism combines information from both camera perspectives using learned attention weights, allowing the model to decide which view is more important for each scene.

The transformer architecture includes positional encoding for temporal sequences and weather-aware attention mechanisms. This allows the model to understand how air quality changes over time and how weather conditions affect the relationship between visual cues and air quality. The multi-task learning framework predicts time-of-day and season alongside the main AQI classification task, which helps the model learn better feature representations.

The ensemble strategy combines predictions from three different transformer models with varying configurations. This reduces prediction variance and improves performance across different environments by leveraging the strengths of each individual model.

\subsection{Feature Extraction and Representation}

Visual feature extraction forms the foundation of our approach using ResNet50 networks. We chose ResNet50 for its proven performance in feature extraction and strong representational capabilities. The feature extraction process works on both front and rear traffic images:

$$\mathbf{F}_f = \Phi_f(I_f) \in \mathbb{R}^{d_f}$$
$$\mathbf{F}_r = \Phi_r(I_r) \in \mathbb{R}^{d_f}$$

\noindent where $\Phi_f$ and $\Phi_r$ represent the front and rear feature extraction networks, and $d_f = 2048$ is the feature dimension from ResNet50's final pooling layer. The networks start with ImageNet pre-trained weights and are fine-tuned on our air quality classification task.

Weather parameters are processed through a standardization and encoding pipeline before being directly integrated with visual features. The numerical meteorological parameters (temperature and humidity) are normalized using standard scaling, while categorical parameters (season and day/night) are encoded as integers. These four processed values are then directly concatenated with the extracted visual features:

$$\mathbf{F}_{combined} = [\mathbf{F}_f; \mathbf{F}_r; W_{norm}] \in \mathbb{R}^{4100}$$

\noindent where $W_{norm}$ represents the normalized and encoded weather parameters, and the semicolon denotes concatenation. The resulting 4100-dimensional feature vector (2048 + 2048 + 4) combines visual information from both camera views with meteorological context, creating a comprehensive representation that captures both the visual characteristics of traffic scenes and the atmospheric conditions that influence air quality.

\subsection{Dual-View Integration Mechanism}

The dual-view integration mechanism is a key part of our approach that combines information from front and rear traffic views. The mechanism uses a learned fusion network with attention-based weighting to adjust the contribution of each view based on the current scene and environmental conditions.

The fusion process starts by computing attention weights that determine how important each view is for the current scene:

$$\alpha_f, \alpha_r = \text{softmax}(W_{\text{fusion}}[\mathbf{F}_f; \mathbf{F}_r])$$

\noindent where $W_{\text{fusion}} \in \mathbb{R}^{2d_f \times 2}$ is a learned parameter matrix that maps the combined features to attention weights. The softmax ensures the attention weights sum to one, creating a weighted combination of the two views:

$$\mathbf{F}_{\text{fused}} = \alpha_f \mathbf{F}_f + \alpha_r \mathbf{F}_r$$

This approach lets the model automatically learn which view provides more useful information for different types of scenes and environmental conditions, rather than using fixed combination rules that might not work well across different scenarios.

\subsection{Enhanced Transformer Architecture}

The transformer architecture includes several components designed for the air quality classification task. The architecture builds on the standard transformer encoder while adding mechanisms to handle the challenges of environmental sensing.

\subsubsection{Temporal Positional Encoding}

To model the sequential nature of air quality data, we use temporal positional encoding that gives the transformer information about the relative positions of samples within time sequences:

$$PE(pos, 2i) = \sin(pos/10000^{2i/d_{model}})$$
$$PE(pos, 2i+1) = \cos(pos/10000^{2i/d_{model}})$$

\noindent where $pos$ is the position within the sequence, $i$ is the dimension index, and $d_{model} = 256$ is the model dimension. This encoding helps the model understand temporal relationships and use the natural patterns in how air quality changes over time.

\subsubsection{Weather-Aware Attention Mechanism}

An important part of our architecture is the weather-aware attention mechanism, which models how meteorological parameters influence the relationship between visual features and air quality conditions. This addresses the challenge that the same visual scene can correspond to different air quality levels depending on atmospheric conditions like temperature, humidity, and season.

The weather-aware attention computes attention weights that adjust the transformer's focus based on current meteorological conditions:

$$\mathbf{A}_w = \sigma(\text{MLP}_{weather}(\mathbf{X} \oplus \mathbf{F}_w))$$
$$\mathbf{X}' = \mathbf{X} \odot \mathbf{A}_w$$

\noindent where $\oplus$ denotes concatenation, $\odot$ represents element-wise multiplication, and $\sigma$ is the sigmoid activation function. This mechanism allows the model to adjust its attention patterns based on weather conditions, improving its ability to generalize across different environmental scenarios.

\subsubsection{Multi-Head Self-Attention Enhancement}

The multi-head self-attention mechanism captures complex relationships between different aspects of the input features. The attention mechanism works as follows:

$$\text{MHSA}(\mathbf{Q}, \mathbf{K}, \mathbf{V}) = \text{Concat}(head_1, ..., head_h)\mathbf{W}^O$$

\noindent where each attention head is computed as:

$$head_i = \text{Attention}(\mathbf{Q}\mathbf{W}_i^Q, \mathbf{K}\mathbf{W}_i^K, \mathbf{V}\mathbf{W}_i^V)$$

$$\text{Attention}(\mathbf{Q}, \mathbf{K}, \mathbf{V}) = \text{softmax}\left(\frac{\mathbf{Q}\mathbf{K}^T}{\sqrt{d_k}}\right)\mathbf{V}$$

\noindent Using multiple attention heads (h = 8 in our implementation) lets the model attend to different aspects of the input features at the same time, capturing both local and global relationships within the data.

\subsection{Multi-Task Learning Framework}

The multi-task learning framework uses the correlations between temporal patterns and air quality dynamics to improve the primary classification task. The framework optimizes three objectives at the same time: primary AQI classification and two auxiliary tasks related to temporal patterns.

The loss function combines these objectives with carefully tuned weights:

$$\mathcal{L}_{total} = \lambda_1 \mathcal{L}_{AQI} + \lambda_2 \mathcal{L}_{time} + \lambda_3 \mathcal{L}_{season}$$

\noindent where:
$$\mathcal{L}_{AQI} = \text{CrossEntropy}(\mathbf{y}_{AQI}, \hat{\mathbf{y}}_{AQI})$$
$$\mathcal{L}_{time} = \text{CrossEntropy}(\mathbf{y}_{time}, \hat{\mathbf{y}}_{time})$$
$$\mathcal{L}_{season} = \text{CrossEntropy}(\mathbf{y}_{season}, \hat{\mathbf{y}}_{season})$$

The weights are set to $\lambda_1 = 1.0$, $\lambda_2 = 0.3$, and $\lambda_3 = 0.3$, giving primary focus to the AQI classification task while using the auxiliary tasks to improve feature quality and model robustness.

% \subsection{Ensemble Strategy and Architecture Diversity}

% The ensemble strategy combines three transformer variants with different configurations to improve robustness and generalization performance. The diversity in the ensemble comes from variations in model architecture, initialization, and training procedures.

% The three models in the ensemble are configured as follows: Model 1 uses a transformer with 8 attention heads, 6 layers, and $d_{model} = 256$; Model 2 has the same configuration as Model 1 but with different random initialization to promote diversity; Model 3 uses a lighter transformer with 4 attention heads, 4 layers, and $d_{model} = 128$ to capture different aspects of the data.

% The final prediction is computed through a learned weighted combination of individual model outputs:

% $$\mathbf{y}_{AQI} = \sum_{i=1}^{3} w_i \mathbf{y}_i^{AQI}$$

% where $w_i$ are learned weights normalized through softmax to ensure they sum to one. This approach allows the model to automatically determine the optimal combination of individual model predictions based on the training data.
% ############################################################################################################################################################################

\section{Experimental Setup}

This section details the datasets, implementation configuration, and evaluation methodology used to assess AQIFormer's performance.
\subsection{Dataset Description and Characteristics}

\subsubsection{TRAQID Dataset}

Our TRAQID dataset~\cite{kathalkar2024traqid} serves as the primary training and evaluation resource for our methodology, comprising 26,678 synchronized front and rear traffic image pairs captured in the urban environment of Hyderabad, India. This dataset represents a comprehensive sampling of diverse environmental conditions, seasonal variations, and traffic scenarios that are representative of Indian urban environments. The data collection was conducted over an extended period spanning multiple seasons, ensuring robust representation of the temporal dynamics that influence air quality patterns.

The temporal distribution of the dataset includes 13,789 daytime images captured during the period from 6 AM to 6 PM, and 12,889 nighttime images captured from 6 PM to 6 AM. This balanced temporal distribution enables comprehensive evaluation of model performance across different lighting conditions, which is crucial for practical deployment scenarios where monitoring systems must operate continuously throughout the day and night.

The seasonal coverage encompasses three distinct periods that characterize the Indian climate: Monsoon season (represented by data from October 2022 and July 2024), Winter season (including January, February, and December 2023), and Summer season (March 2024). This comprehensive seasonal representation ensures that the model can generalize across the significant environmental variations that occur throughout the year in Indian cities.

Each image pair is associated with comprehensive metadata including precise timestamps, GPS coordinates, and co-located environmental measurements. The environmental data encompasses temperature measurements, relative humidity readings, PM2.5 and PM10 concentrations, and computed AQI values according to CPCB standards. This rich metadata enables detailed analysis of the relationships between environmental conditions and visual air quality indicators.

\begin{figure}[!b]
  \centering
  \includegraphics[width=\columnwidth]{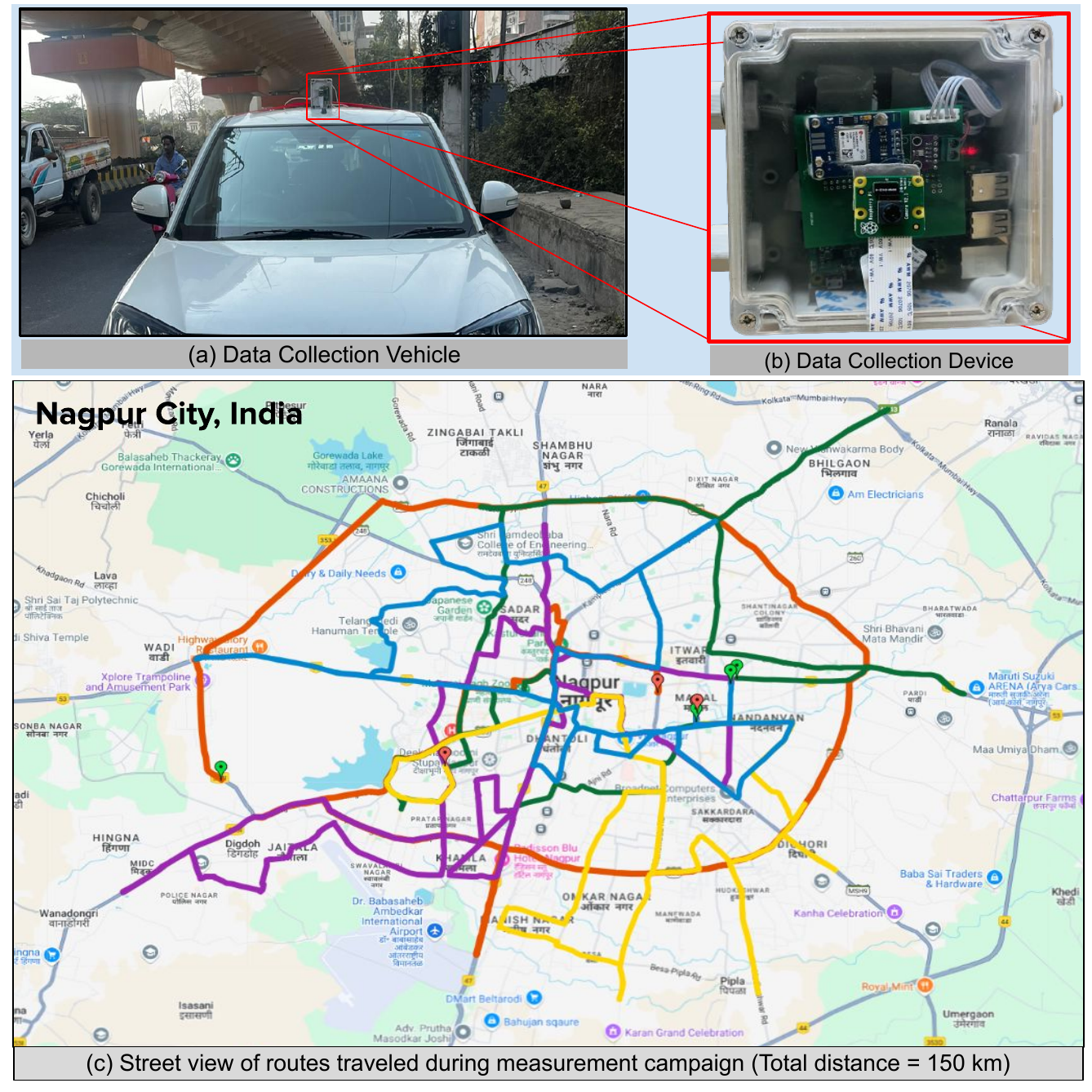}
  \caption{Nagpur dataset collection setup showing (a) data collection vehicle, (b) IoT sensing device with RPi4 Camera and environmental sensors, and (c) systematic route map showing comprehensive spatial coverage (150Km).}
  \label{fig:map_nagpur}
  \Description{Route Map Nagpur}
\end{figure}

\subsubsection{Nagpur Dataset}

To evaluate cross-city generalisation capabilities, we collected an independent validation dataset in Nagpur, Maharashtra, India. Inspired by established traffic-based air quality monitoring approaches, this dataset tests the model's ability to generalize across different urban environments.

Data collection was carried out between January 16-20, using a Raspberry Pi 4 (RPi4) microcontroller interfaced with a PiCamera for traffic image capture. Environmental sensors included a BME280 for temperature and humidity measurements, and a Nova PM SDS011 sensor for PM2.5 and PM10 concentrations. The Nova SDS011 was selected for its proven reliability in PM detection. AQI values were calculated using the standard CPCB formulation based on measured PM concentrations, which serve as ground-truth labels.

The collection followed a systematic route-based approach covering major traffic corridors in Nagpur city (Figure~\ref{fig:map_nagpur}). Data acquisition occurred at 30-second intervals during both peak and off-peak traffic periods, encompassing diverse traffic densities and urban environments. The Nagpur data set comprises 2,371 front-view traffic images with synchronised environmental metadata collected over 4 days. All samples represent daytime conditions (6 am - 6 pm) during the winter season. The environmental parameters include temperature (28.1 ° C - 41.1 ° C), relative humidity (25.9\% - 42.7\%) and precise timestamps with temporal categorisation.

The AQI distribution reflects distinct pollution dynamics: Satisfactory (0.8\%), Moderate (32.3\%), Poor (50.3\%), Very Poor (15.6\%), and Severe (0.9\%). In particular, no samples fall into the Good category, indicating consistently elevated levels of pollution during the collection period. Nagpur's compact urban centre with radial road networks and higher commercial vehicle concentrations due to its logistics hub role provides distinct characteristics for testing generalisation capabilities. The data set enables the evaluation of zero-shot generalization and domain adaptation in different urban environments.

\subsection{Implementation Details and Training Configuration}

The implementation of AQIFormer utilizes PyTorch~\cite{pytorch2019} as the primary deep learning framework, leveraging its flexibility and comprehensive ecosystem for transformer-based architectures. The ResNet50 backbones are initialized with pre-trained weights from ImageNet to leverage transfer learning capabilities while being fine-tuned for the specific air quality classification task. The transformer encoder architecture consists of 6 layers with 8 attention heads and 256-dimensional features, providing sufficient capacity to model complex spatio-temporal relationships while maintaining computational efficiency.

The training procedure employs the AdamW optimiser with an initial learning rate of 1e-4 and weight decay of 1e-5 to prevent overfitting. The learning rate schedule utilizes cosine annealing to gradually reduce the learning rate throughout training, promoting better convergence and final performance. Mixed-precision training is used to improve computational efficiency and reduce memory requirements while maintaining numerical stability.

All experiments were conducted on dual NVIDIA GeForce GTX 1080 Ti GPUs~\cite{nvidia2017} with 11~GB VRAM each, utilizing CUDA 12.8 for accelerated training. The distributed training setup enables efficient processing of large batch sizes while maintaining memory constraints within acceptable limits. Standard data augmentation techniques including random horizontal flips and color jittering are applied to enhance model robustness while preserving air quality-relevant visual characteristics.

For cross-city evaluation, few-shot adaptation was performed using approximately 120 samples (5\% of the Nagpur dataset) for fine-tuning the pre-trained Hyderabad model. The adaptation process involved minimal architectural modification to handle single front-view images and brief fine-tuning (3 epochs) with a reduced learning rate of 1e-5 to preserve the learned representations while adapting to local characteristics. Early stopping is implemented with a patience of 7 epochs based on validation loss to prevent overfitting and ensure optimal generalization performance.

\subsection{Evaluation Metrics and Performance Assessment}

The evaluation of AQIFormer employs multiple complementary metrics to provide comprehensive assessment of model performance across different aspects of the air quality classification task. Primary metrics include classification accuracy, which measures the proportion of correctly classified samples across all AQI categories, and weighted F1-score, which provides a balanced measure of precision and recall weighted by the number of samples in each class to account for class imbalance.

Additional evaluation includes detailed confusion matrix analysis to understand class-specific performance and identify potential systematic biases in the model's predictions. Precision and recall are computed for each AQI category to provide insights into the model's ability to correctly identify each level of air quality. Cross-city generalisation is assessed through few-shot adaptation scenarios where models trained on one city are fine-tuned on a small subset of data from another city, demonstrating practical deployment feasibility with minimal data requirements.

The evaluation framework also includes ablation studies to understand the contribution of individual architectural components, seasonal and temporal performance analysis to assess robustness across different environmental conditions, and computational efficiency metrics to evaluate the practical feasibility of deployment in resource-constrained environments.

\section{Results and Analysis}

This section presents comprehensive experimental results demonstrating AQIFormer's performance across within-city evaluation, cross-city generalization, and detailed ablation studies.

\subsection{Within-City Performance Evaluation}

The within-city evaluation of AQIFormer on the TRAQID dataset demonstrates significant improvements over existing state-of-the-art methods for image-based air quality classification. Our comprehensive evaluation reveals that AQIFormer achieves 89.96\% accuracy with a weighted F1-score of 0.88, representing substantial improvements over all baseline methods evaluated.

\begin{table}[!t]
\centering
\caption{Within-city AQI classification results on TRAQID dataset}
\label{tab:within_city}
\begin{tabular}{@{}lcc@{}}
\toprule
Method & Accuracy (\%) & F1-Score \\
\midrule
Mondal et al.~\cite{mondal2024uncovering} & 64.0 & 0.61 \\
Kalajdjieski et al.~\cite{kalajdjieski2020air} & 60.0 & 0.56 \\
Nilesh et al.~\cite{nilesh2022iot} & 73.0 & 0.71 \\
AQC-Net~\cite{zhang2020deep} & 75.0 & 0.74 \\
\midrule
\textbf{AQIFormer (Ours)} & \textbf{89.96} & \textbf{0.88} \\
\bottomrule
\end{tabular}
\end{table}

\begin{figure*}
\centering
\includegraphics[width=\textwidth]{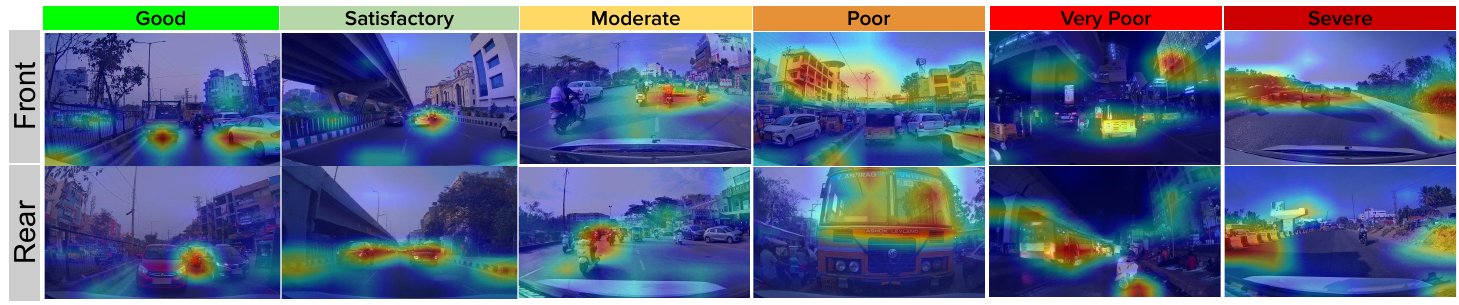}
\caption{Attention visualization maps across AQI categories showing AQIFormer's focus regions for front and rear traffic views.}
\label{fig:attention_maps}
\Description{TRAQID dataset attention maps}
\end{figure*}

\begin{figure*}
\centering
\includegraphics[width=\textwidth]{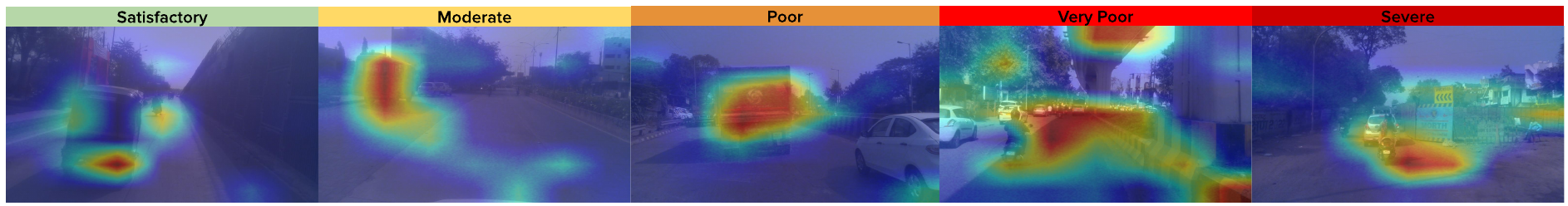}
\caption{Cross-city attention visualization on Nagpur dataset showing consistent attention patterns: localized vehicle focus in Satisfactory conditions, traffic areas in Moderate conditions, dense hotspots in Poor/Very Poor conditions, and multi-source attention in Severe conditions.}
\label{fig:attention_maps_nagpur}
\Description{TRAQID dataset attention maps}
\end{figure*}

The results demonstrate that AQIFormer achieves a 14.96\% improvement in accuracy compared to the best-performing baseline method (AQC-Net) and an 18.9\% improvement in F1-score. These improvements are particularly significant considering the challenging nature of the dataset, which includes diverse environmental conditions, seasonal variations, and complex traffic scenarios that represent real-world deployment challenges.

The good performance of AQIFormer can be attributed to several key architectural factors. The dual-view integration mechanism enables the model to capture complementary information from both front and rear traffic perspectives, providing a more comprehensive understanding of the traffic scene and atmospheric conditions. The enhanced transformer architecture with weather-aware attention mechanisms allows the model to adaptively process visual features based on meteorological conditions, improving its ability to handle the complex relationships between environmental factors and air quality indicators. Notably, AQIFormer represents the first successful approach to achieve reliable nighttime AQI classification from traffic imagery, maintaining only 1.68\% performance degradation under challenging low-light conditions.

% After the existing content in Section 4.1
The attention visualization analysis reveals that AQIFormer learns meaningful category-specific focus patterns that demonstrate both interpretability and robust generalization capabilities. As shown in Figure~\ref{fig:attention_maps}, the model exhibits distinct attention behaviors across AQI categories: for Good conditions, attention focuses on individual vehicle exhaust regions; Satisfactory/Moderate categories show expanded focus on traffic intersections and congestion zones; Poor/Very Poor conditions concentrate attention on heavy traffic hotspots and dense vehicular clusters; while Severe conditions demonstrate sophisticated multi-source attention encompassing traffic congestion, construction activities, and urban infrastructure elements. Remarkably, these attention patterns remain consistent across different urban environments, as demonstrated by the Nagpur cross-city evaluation (Figure~\ref{fig:attention_maps_nagpur}), where the model maintains similar category-specific focus strategies despite never being trained on Nagpur data. This consistency validates both the model's ability to identify physically meaningful pollution sources and its robust generalization capabilities across diverse urban contexts.

\subsection{Cross-City Generalization Performance}

The cross-city generalization evaluation represents one of the most significant contributions of this work, addressing a critical challenge in the practical deployment of image-based air quality monitoring systems. The evaluation involves training AQIFormer on the Hyderabad dataset and testing it on the Nagpur dataset using few-shot adaptation with a small subset of Nagpur training samples (approximately 5\% of the dataset) to accommodate the single front-view architecture and local environmental characteristics.

\begin{table}[!t]
\centering
\caption{Cross-city generalization results (Hyderabad → Nagpur)}
\label{tab:cross_city}
\begin{tabular}{@{}lcc@{}}
\toprule
Method & Accuracy (\%) & F1-Score \\
\midrule
AQC-Net~\cite{zhang2020deep} & 69.0 & 0.67 \\
Nilesh et al.~\cite{nilesh2022iot} & 73.0 & 0.71 \\
\midrule
\textbf{AQIFormer (Ours)} & \textbf{81.67} & \textbf{0.79} \\
\bottomrule
\end{tabular}
\end{table}

The cross-city validation results demonstrate exceptional generalization capabilities, with AQIFormer maintaining 81.67\% accuracy and 0.79 F1-score when applied to the Nagpur dataset. The performance degradation compared to within-city evaluation is only 8.29\% for accuracy, which represents remarkably strong generalization considering the significant differences between the two cities in terms of urban layout, traffic patterns, and environmental conditions. This level of cross-city generalization significantly outperforms existing methods, with baseline approaches showing much larger performance degradations when applied to new urban environments. This generalization capability of AQIFormer suggests that the model learns fundamental relationships between visual characteristics and air quality rather than city-specific patterns, making it highly suitable for practical deployment in diverse urban environments.

\subsection{Comprehensive Ablation Studies}

To understand the contribution of each architectural component and validate the design choices in AQIFormer, we conducted comprehensive ablation studies that systematically evaluate the impact of different components on overall performance.

\begin{table}[!t]
\centering
\caption{Ablation study results on TRAQID dataset}
\label{tab:ablation}
\begin{tabular}{@{}lcc@{}}
\toprule
Model Variant & Accuracy (\%) & F1-Score \\
\midrule
\textbf{Complete Model} & \textbf{89.96} & \textbf{0.88} \\
\midrule
Front Only & 78.50 & 0.75 \\
Rear Only & 71.55 & 0.70 \\
No Weather Data & 68.52 & 0.65 \\
No Categorical Features & 78.39 & 0.78 \\
Image Only & 61.42 & 0.58 \\
\bottomrule
\end{tabular}
\end{table}

The ablation study results provide several key insights into the importance of different architectural components. The dual-view integration mechanism demonstrates significant value, with the complete model achieving 19.46\% higher accuracy than the best single-view approach (front-only). This improvement validates the hypothesis that front and rear views provide complementary information that enhances air quality classification performance.

The integration of weather data shows a substantial impact, with the complete model achieving higher accuracy than the image-only baseline. This improvement demonstrates the critical importance of incorporating meteorological information for accurate air quality assessment, as the same visual scene can correspond to different air quality levels depending on atmospheric conditions.

The categorical features including season and time-of-day information contribute 11.57\% improvement in accuracy, highlighting the importance of temporal context in air quality assessment. The multi-task learning framework that incorporates these auxiliary tasks provides additional supervision signals that improve feature quality and model robustness.

\subsection{Seasonal and Temporal Performance Analysis}

\begin{table}[!t]
\centering
\caption{Seasonal and temporal performance analysis}
\label{tab:seasonal}
\begin{tabular}{@{}lcc@{}}
\toprule
Condition & Accuracy (\%) & Characteristics \\
\midrule
\textbf{Winter} & \textbf{91.24} & Optimal atmospheric conditions \\
Monsoon & 88.72 & Moderate weather variations \\
Summer & 85.44 & Challenging extreme conditions \\
\midrule
\textbf{Day} & \textbf{89.84} & Better lighting conditions \\
Night & 88.16 & Slight illumination degradation \\
\bottomrule
\end{tabular}
\end{table}

The analysis of seasonal and temporal performance variations provides insights into the model's robustness across different environmental conditions and its ability to capture the complex dynamics of air quality patterns. The seasonal analysis reveals that AQIFormer achieves the highest accuracy during winter conditions (91.24\%) when atmospheric conditions are typically more stable and conducive to accurate visual air quality assessment. The model maintains strong performance during the monsoon season (88.72\%) despite the challenges posed by variable weather conditions, including rain and high humidity. Summer conditions present the greatest challenge (85.44\%) due to extreme temperatures, high solar radiation, and complex atmospheric dynamics, yet the model still maintains robust performance.

The temporal analysis shows that daytime performance (89.84\%) slightly exceeds nighttime performance (88.16\%), with only 1.68\% degradation during nighttime conditions. This minimal performance difference demonstrates the effectiveness of the model's ability to handle illumination variations and extract meaningful air quality information from images captured under different lighting conditions.

\section{Conclusion}

This paper presents AQIFormer, a transformer-based ensemble architecture that addresses fundamental challenges in image-based air quality classification through dual-view integration, weather-aware attention mechanisms, and multi-task learning. The extensive experimental evaluation demonstrates significant improvements over existing state-of-the-art methods, achieving 89.96\% accuracy on our TRAQID dataset and maintaining 81.67\% accuracy in cross-city generalization scenarios.

The comprehensive ablation studies validate the contribution of each architectural component, with dual-view integration providing 11.46\% improvement over single-view approaches, followed by multi-task learning contributing 11.57\% enhancement. The cross-city generalization capabilities, with only 8.29\% performance degradation when transferring from Hyderabad to Nagpur using few-shot adaptation, establish AQIFormer as a practical solution for scalable urban air quality monitoring.

AQIFormer's robust performance and cross-city generalization enable practical deployment for cost-effective air quality monitoring using existing traffic camera infrastructure. The 81.67\% cross-city accuracy achieved through few-shot adaptation demonstrates scalability across diverse urban environments with minimal data and computational requirements for local adaptation. The generalisation capabilities across different urban environments make this approach highly suitable for practical deployment in diverse Indian cities. To support the research community and enable further advancement in cross-city air quality monitoring, we will release the Nagpur dataset as an open resource.

However, several limitations warrant consideration: nighttime performance shows 1.68\% degradation partially due to headlight glare effects that can interfere with visual feature extraction, and evaluation remains limited to two Indian cities requiring validation across broader geographic regions and extreme weather conditions such as heavy fog or dust storms. The transformer architecture's computational requirements may limit real-time edge deployment, suggesting future research directions in model compression and lightweight architectures for resource-constrained environments. Future research directions include extending evaluation to more diverse geographic regions, developing lightweight architectures for edge deployment, and exploring federated learning approaches for privacy-preserving multi-city training. The foundation established by AQIFormer provides a platform for continued advancement in image-based environmental monitoring and smart city applications.
\begin{acks}
We thank iHubData, IIIT Hyderabad for extending research fellowship and Bodhyan platform support for curating the novel dataset.
\end{acks}

\bibliographystyle{ACM-Reference-Format}
\bibliography{references}

\end{document}